\documentclass[a4paper,10pt]{llncs}

\usepackage{graphicx}
\usepackage{comment}
\usepackage{booktabs}
\usepackage{tabu}
\usepackage{wasysym}
\usepackage{color,soul}
\graphicspath{{./pics/}}
\usepackage{multirow}
\usepackage{hhline}
\let\oldfootnote\footnote
\def\footnote{\ifhmode\unskip\fi\oldfootnote}

\begin{document}

\author{Alireza Sedghi \inst{1,2}
\and Jie Luo\inst{2,3}
\and Alireza Mehrtash \inst{2,4} 
\and Steve Pieper \inst{2} 
\and Clare M. Tempany \inst{2}
\and Tina Kapur \inst{2} 
\and Parvin Mousavi \inst{1}
\and William M. Wells III \inst{2} 
}

\institute{
Medical Informatics Laboratory, Queen’s University, Kingston, ON, Canada \\
\email{sedghi@cs.queensu.ca}\\
\and Radiology Department, Brigham and Women’s Hospital, Harvard Medical School, Boston, MA, USA \\
\and
Graduate School of Frontier Sciences, The University of Tokyo, Japan
\and Department of Electrical and Computer Engineering, University of British Columbia, Vancouver, BC, Canada
}

\title{Semi-Supervised Deep Metrics for Image Registration}
\titlerunning{Semi-Supervised Deep Metrics for Image Registration}

\maketitle

\begin{abstract}
Deep metrics have been shown effective as similarity measures in multi-modal image registration; however, the metrics are currently constructed from aligned image pairs in the training data.  In this paper, we propose a strategy for learning such metrics from roughly aligned training data. Symmetrizing the data corrects bias in the metric that results from misalignment in the data (at the expense of increased variance), while random perturbations to the data, i.e. {\em dithering}, ensures that the metric has a single mode, and is amenable to registration by optimization. Evaluation is performed on the task of registration on separate unseen test image pairs. The results demonstrate the feasibility of learning a useful deep metric from substantially misaligned training data, in some cases the results are significantly better than from 
Mutual Information. Data augmentation via dithering is, therefore, an effective strategy for discharging the need for well-aligned training data; this brings deep metric registration from the realm of supervised to semi-supervised machine learning.\\

\textbf{Keywords}: multi-modal image registration, deep learning, classification
\end{abstract}

\section{Introduction}

While deep learning-based metrics \cite{simonovsky2016deep,cheng2016deep} demonstrate great promise for solving difficult registration problems, they are thought
to require well-registered training data, which can be a serious drawback.
Consider the registration of abdominal MRI and Ultrasound (US), an important and difficult problem.
There
is strong application pull for such registration, as diagnostic MRI has good contrast for tumors, while
US is routinely used for interventional guidance.
Unfortunately, for this application, it is difficult to obtain well-registered training data,
because
the scans can not be obtained simultaneously, and the anatomy will shift between scans. While manual registration is a possibility,
it is laborious and technically challenging.  For these reasons, there is great practical advantage if we can 
eliminate or reduce the requirement for well-registered training data for learning deep metrics for registration.
In this paper we show how this may be achieved.

Many registration systems can be decomposed into two components, an image agreement metric (or objective function), and an optimizer.
Human designed agreement metrics such as mutual information (MI) have been successful in multimodal image registration \cite{viola1997alignment}. However, for difficult problems, single pixel statistics may not capture all the information that is 
needed. In addition, using manually constructed features also limits the capacity to learn  the  information that is shared among
images.

Convolutional Neural Networks (CNNs) have proven remarkably powerful for image classification, and other image processing tasks, 
presumably because they are able to learn and manipulate effective representations of image contents at multiple levels of abstraction.
They are now gaining traction on difficult medical imaging problems.
A recent survey on deep learning methods in medical image analysis \cite{Litjens2017ASO} reports two common themes where deep neural networks have been applied to image registration: 1) estimation of similarity measures \cite{simonovsky2016deep,Wu2013UnsupervisedDF,cheng2016deep}; 2) estimation of transformation parameters, between the images  \cite{Sokooti2017NonrigidIR,Miao2016ACR}. Here, we will focus on the first category as our work is also estimation of similarity metrics. Wu \textit{et al.} \cite{Wu2013UnsupervisedDF} explored unsupervised learning methods to extract deep features from the input patches and used the learned features vectors, instead of hand-crafted features, in an existing registration method. Cheng \textit{et al.} \cite{cheng2016deep} proposed learning such a metric by training on corresponding and non-corresponding patches from CT and MR in a multi-modal stacked denoising autoencoder framework. Simonovsky \textit{et al.} \cite{simonovsky2016deep}, trained a CNN classifier to distinguish between pairs of corresponding and non-coresponding patches. After training, gradients of the deep metric were used to compute the updates for the transformation parameters in an iterative manner. A limitation of this approach appears to be that well-registered training data is a requirement for training such a classifier. 

In this paper we focus on the training requirements of deep metrics for registration.
We demonstrate that well-registered training data is actually not required. With misregistered training data there is a risk that the objective function will be biased, however, we demonstrate that with  suitable data augmentation, 
including a novel ``dithering" approach, the effect of misregistration is to broaden the objective function, while eliminating its bias, in comparison to the non-augmented case. While there may be some loss of accuracy due to the broadening of the objective function, we show that a multi-shot approach can be used whereby the broadened response function is used to improve the registration of the training data, and the process repeated. This leads to a well-registered training data set where the ultimate trained network would perform as well as the one presented in \cite{simonovsky2016deep}.   We envision that our training approach will be used
once for a new application domain, subsequently registrations in the domain will only require the trained network.


\begin{figure}[ht!]
\centering
\vspace{0cm}
{\includegraphics[scale=1.1 ]{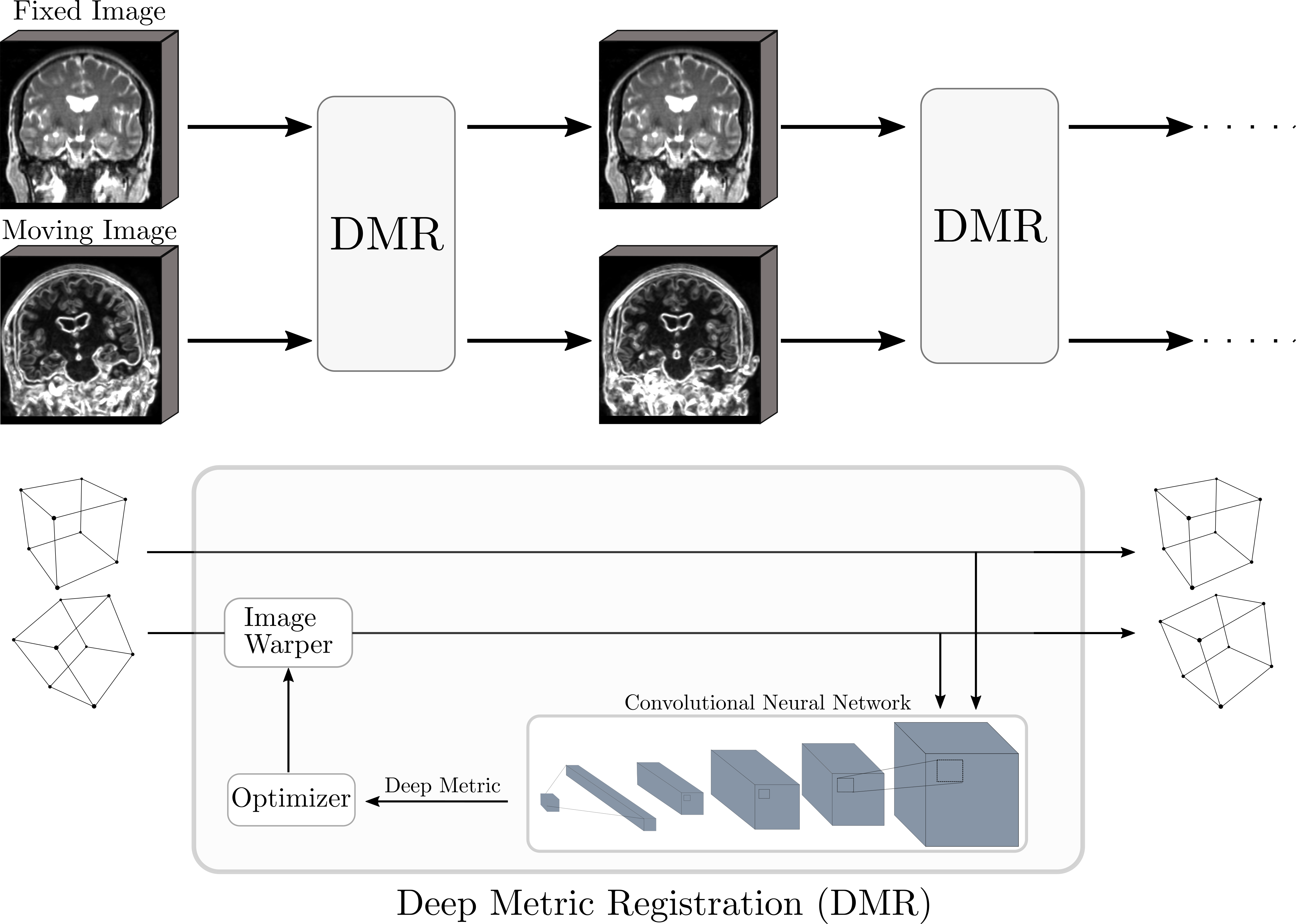}}
\caption{Overview of our training method for learning a deep metric from misaligned data (data augmentation is not shown in the figure).}
\label{fig:diagram}
\end{figure}

\section{Method}

Registration between a fixed image $u(x)$ and a moving image $v(x)$ can be formulated as optimizing a transformation $\mathcal{T_\theta}$, in transformation space $T$, that maximizes their agreement measured by the (similarity) score $F$.

\begin{equation}
   \hat{\mathcal{T}_\theta} = arg \max\limits_{\theta} \: F(u(x), v(\mathcal{T_\theta}(x)).
\end{equation}

In this context, $\mathcal{T_\theta}$ is a transformation from the coordinate frame of the moving
image to the fixed image, $x$ are locations in the  coordinate frame, and $\theta$ are the
transformation parameters.
Mutual information compares images by measuring the statistical relationship among single pixels of two modalities; however, deep networks compare
images using a hierarchy of concepts (edges, contours, etc.) from all the pixels, and we hypothesize this will substantially improve the performance in non-trivial registration problems. In this work, we estimate a similarity metric ($F$) from the output score of classifier CNNs by training over 
corresponding and non-corresponding pairs of patches. In our framework we learn the metric from roughly aligned data, in contrast to the method in
\cite{simonovsky2016deep} where perfectly aligned data were used for the training. By applying symmetrization and dithering to the training data, we demonstrate that even with substantial misalignment between the modalities, a deep metric can be learned and further used for the task of registration. Our metric is formulated as the aggregation of the score of the network, $M$, for $n$ randomly chosen patches, $p$, from the fixed and moving image:
\begin{equation}
    F(u(x),v(\mathcal{T_\theta}(x)))=\sum_{n} M(u(p),v(\mathcal{T_\theta}(p)))
\end{equation}
In our registration experiments we optimize the metric by Powell's method, a popular non-derivative optimizer \cite{powel}.

\subsection{Data Augmentation by Symmetrization and Dithering }\label{dither}
Depending on the distribution of misregistration in the training data, 
without any augmentations applied, there may be substantial bias in the deep metric response function. For instance, if the moving images in the training data were all shifted $10~mm$ to the right, then the response function will reflect this bias with a peak at $10~mm$. 
Such bias can be effectively addressed by symmetrizing the 
training data; in the above example, flipping half of the moving images in the training
set about the vertical axis results in a response function that now has two prominent modes at 
$\pm 10~mm$. Although the bias has
been greatly reduced, the bimodal property of the response function will
cause major problems for optimizers. Historically, Gaussian blur of the images has been used to achieve a single mode in objective functions, leading to  multi-resolution
approaches \cite{650848}. We experimented with such blurring; however, we found
that ``dithering" is a more effective approach. In signal processing, dithering
is the deliberate introduction of noise that can be used for randomizing quantization error \cite{1088973}.
We apply Gaussian distributed random displacements to the moving image to effectively 
merge modes whose separation is on the order of the standard deviation of the dithering, as follows,
$v(x_i) \rightarrow v(x_i+d)$ where $d \sim N(0,\sigma^2I)$ and $\sigma^2$ is variance of the dither. Pairs of 3D patches are randomly cropped from the fixed and their corresponding locations in the moving image. For the non-corresponding class, patches are selected randomly in space. Empirically we observed an increase in the performance by applying both restrictions of position (minimum distance between misregistered patches), and sampling with restriction to the anatomy. Finally, all the patches are augmented by a combination of rotation and flipping to remove the bias as explained earlier. In the situations where we have substantial misregistration in the training data, small patch sizes can not fully capture 
long range shared information between the images. Therefore, we propose to perform multi-resolution, multi-shot registration by applying downsampling to the images. In our method, the learned deep metric in the downsampled data is used to realign the training data which eventually will make the misalignment smaller and enable registration with smaller patch sizes. To accomplish anti-aliasing in the downsampling, a Gaussian kernel with a standard deviation proportional to the level of downsampling were applied to smooth the images, and intensities were normalized to $[0,1]$. On average, 1 million patches were generated for training of each CNN which consists of
$\{ (u_i,v_i,z=1)...(u_j,v_j,z=0)... \}$
where the $u_i$ and $v_i$ are augmented versions of the roughly registered patches from training data and $u_j$ and $v_j$ are uniformly randomly misregistered patches.

\subsection{Network Architecture and Training}
\textbf{Architecture:} A 2-channel 3D CNN was used to estimate the similarity of two 3D volumes. The 2-channel architecture has the capacity to compare the patches from the beginning (early fusion), and has exhibited better performance among all other networks in the task of patch comparison \cite{ZagoruykoK15}. We stacked the fixed image patch ($u(p)$) and moving image patch ($v(p)$) as input channels of the network. Specifically, we used a 5-layer architecture consisting of 3 convolutional layers (filter size 5 followed by 3 layers of filter size 3, all stride 2), a pooling layer, a dense layer and a final softmax layer.

\textbf{Training:} We train our network by minimizing the cross-entropy loss between registered and unregisterd patches. To tune the weights of CNN, the Adam  \cite{Kingma2014AdamAM} update rule with a decaying learning rate of 0.01 (decay factor of 0.8 after each epoch) is used. To prevent overfitting, we also add dropout \cite{Srivastava2014DropoutAS} with a drop rate of 0.5, and regularize the weights with $L_2$ weight decay with $\lambda=  0.01$ penalty  on  the  weights.

\textbf{Registration:}
We use the signal immediately before the final softmax as the per-patch response function because the softmax
tends to flatten the response functions, which can cause difficulties for registration optimizers.
In probabilistic terms, the pre-softmax value corresponds to the log likelihood ratio for the
registered vs. unregisterd classes (rather than the posterior probability of ``registered").
After using dithering to shape the response function, 
we use the trained network's output summed over multiple patches as an objective function to register new data, or to improve registration of the training data,
which will greatly reduce bias, and reduce variance, in the misregistration of the training data.  This process
may be repeated, if necessary as depicted in Figure \ref{fig:diagram}.


\section{Experiments and Results}

\textbf{Materials:} We use IXI brain development dataset \cite{BrainDev33:online} which consists of approximately 600 aligned T1 and T2-weighted MRI, PD, MRA and DTI scans of adult brains. In the first two experiments, we randomly choose 100 pairs of T1, T2-weighted scans, and use 25 pairs for training and validation, and the rest for the evaluation. All the images are resampled to $1\times1\times1~mm$. Patch size of $17\times17\times17$ were used for all experiments except experiment 1 in which we used patch size of $32\times32\times5$ for the first part.

\textbf{Experiment1:} We claim that misalignment in the training data can appear as a substantial bias in the deep metric response function, and applying symmetrization will make the shifted response function bimodal. To support this claim, we train separate CNNs on two datasets. In the first, one of the images is deterministically translated by $8mm$ along the $x$ axis. In the second, we generate an augmented version of the previous data set by combinations of rotation, and mirroring applied to the fixed and moving patches. 
The deep metric is characterized by plotting, as a function of, e.g., translation, the sum of pre-softmax activations of the 64 randomly chosen patches in a test data. 

{We study the effect of dithering on a training set 
that has substantial bias in the distribution of misregistration
by rigidly transforming the moving image following a misalignmnet. We apply a 3D uniformly distributed translation $\mathcal{U}_{t}\{1,5\}~mm$ in the $x, y, z$ axis
followed by a 3D uniformly distributed rotation $\mathcal{U}_{\theta}\{-0.1,0.1\}~rad$. We train a CNN on a dithered version of the data by applying $m=100$ 3D random translation following a Gaussian distribution $d \sim N(0,\sigma^2I)$ with $\sigma^2=10$.}

\textbf{Experiment2:} In this experiment we demonstrate the effectiveness of our proposed method in substantially misregistered training data, for intra-subject rigid registration of T1-T2 MRI scans.  A rigid transformation, with translation and rotation parameters sampled from $\mathcal{U}_{t}\{1,20\}\allowbreak~mm$ and $\mathcal{U}_{\theta}\{-0.2,0.2\}~rad$, is applied to misregister the moving images (T1). Our baseline for this experiment is normalized mutual information with 60 histogram bins. Having a large range of misalignment, we experiment with a multi-resolution, multi-shot training strategy. Specifically, a 4 stage training is performed where each stage's learned deep metric is used to realign the training data to reduce the misregistraion. Model selection is performed based on the accuracy on the validation set. We propose 4 stages as $F_1(4,100) \rightarrow F_2(2,25) \rightarrow F_3(2,15) \rightarrow F_4(1,8)$ where $F(l,\sigma^2)$ represents learning a deep metric by downsampling the data by factor of $l$ and dithering with a variance $\sigma^2$. We also  augment each patch by rotation and flipping. We misregister the unseen test set in the same manner as training, and use $F1-F4$ to rigidly register the data. 

\textbf{Experiment3:} To explore the accuracy of the proposed method in handling misalignments on other domains, we perform experiment with learning a deep metric in a dataset of T2-weighted MRI, and Gradient Magnitude (GM) images  (Figure \ref{fig:diagram}) derived from T1 MRI scans; the GM images are meant
to be somewhat similar to US images. We randomly select $k=125$ cases from the IXI dataset and use 75 for training and validation, and the rest for evaluation. After converting all the T1 images to GM images, we misalign the data by a rigid registration following $\mathcal{U}_{t}\{1,10\}~mm$, and $\mathcal{U}_{\theta}\{-0.1,0.1\}~rad$. In this experiment we perform a 3-stage learning strategy, specifically $F_1(2,25) \rightarrow F_2(2,15) \rightarrow F_3(1,5)$ to effectively learn the deep metric from the misaligned training data, and we apply $F_1-F_3$ for registering the misaligned evaluation data.

\begin{figure}[ht!]
\centering
\vspace{0cm}
{\includegraphics[width=4 in]{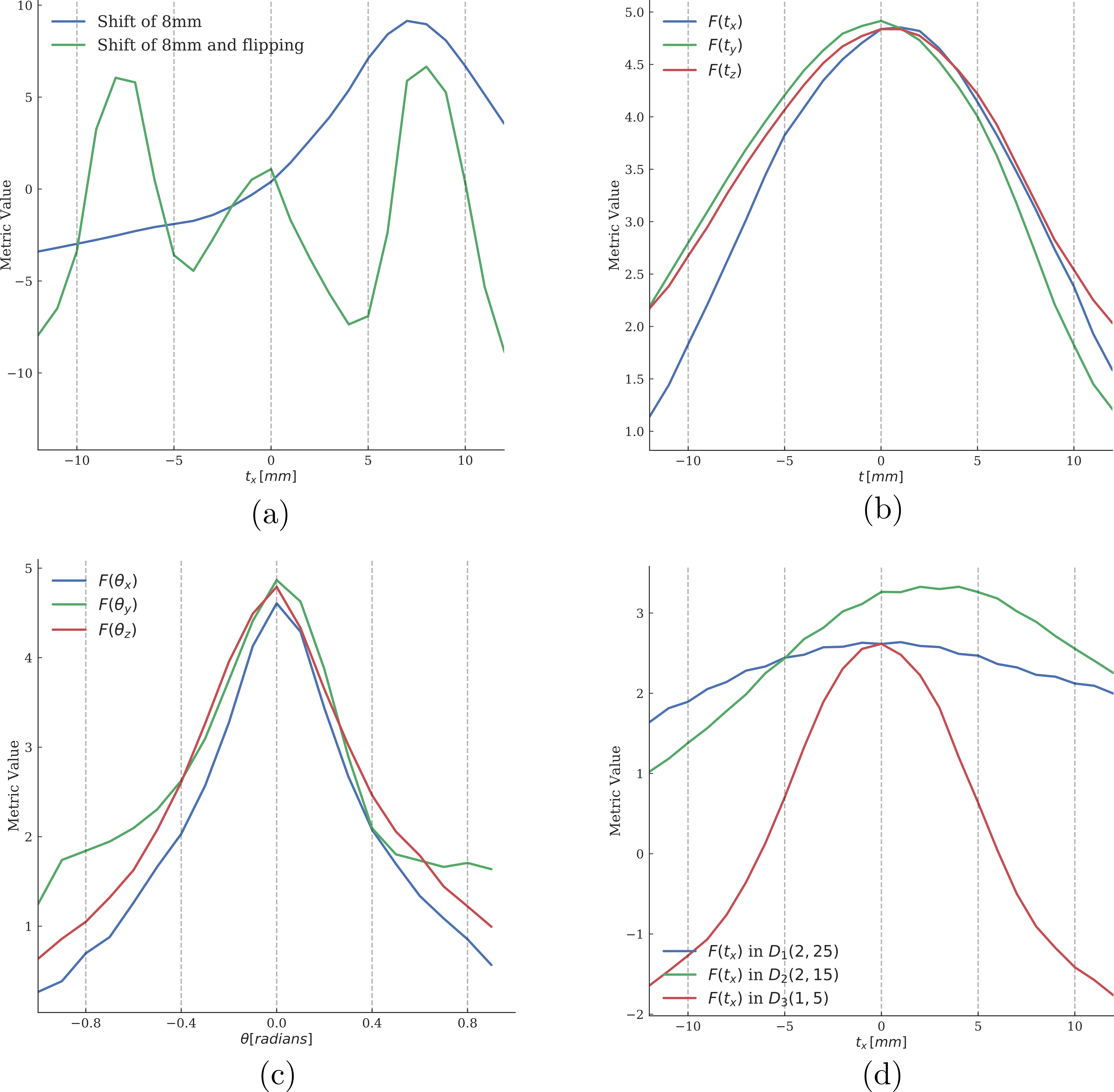}}
\caption{Activation Functions: (a) symmetrization reduces bias, leading to multiple modes, (b) dithering produces a single mode in translation and (c) rotation,
(d) multi-resolution multi-shot experiment: initial response is broad and somewhat biased, intermediate response is broad with less bias, final response is less broad and has little bias. }
\label{fig:activations}
\end{figure}

\textbf{Results:}
Figure \ref{fig:activations}a demonstrates the effect of distribution of misregistration in the training data, on the deep metric response function for experiment 1. In Figures \ref{fig:activations}b and \ref{fig:activations}c we can see the unimodal and broad response function that is caused by dithering of the moving image. The quantitative results of experiment 2 and 3 are listed in Table \ref{results}. For experiment 2, results demonstrates statistically significant improvement of the Euclidean norm of the translation error ($\|T\|$) and the mean absolute error of the translation in $x$ axis ($t_x$) ($p<0.01$) by our proposed method. The 3 stage response functions in the experiment 4 are estimated on a test data as a function of $t_x$ and depicted in the Figure \ref{fig:activations}d. In addition, quantitative results of this experiment show significantly improved performance in the norm of the translation error, $p<0.0001$, compared to MI for T2-GM registration problem

\renewcommand{\arraystretch}{1.5}
\begin{table}[]
\centering
\caption{Quantitative rigid registration results for MI and CNN trained on a misregistered data, unit of translation is $mm$ and unit of the rotation is converted to degrees for easier interpretation}
\label{results}
\resizebox{\columnwidth}{!}{
\begin{tabular}{l|l|c|c|c|c|c|c|c}
\hline
Task & Method & $\|T\|$ & $\overline{|\Delta t_x|}$ & $\overline{|\Delta t_y|}$ & $\overline{|\Delta t_z|}$ & $\overline{|\Delta \theta_x|}$ & $\overline{|\Delta \theta_y|}$ & $\overline{|\Delta \theta_z|}$ \\ \hline
\hline
\multirow{2}{*}{T1-T2} & MI & $4.57 \pm 2.58$ & $2.84 \pm 2.39$ & $1.83 \pm 1.69$ & $2.19 \pm 1.67$ & $1.89 \pm 1.37$ & $1.60 \pm 1.26$ & $1.54 \pm 1.14$ \\ \cline{2-9} 
 & CNN & $ 3.62 \pm 1.68$ & $1.72 \pm 1.46$ & $1.49 \pm 1.50$ & $2.06 \pm 1.45$ & $2.0 \pm 1.43$ & $1.83 \pm 1.71$ & $1.71 \pm 1.48$ \\ \hline
\multirow{2}{*}{T1-GM} & MI & \multicolumn{1}{l|}{$2.02 \pm 0.70$} & \multicolumn{1}{l|}{$1.10 \pm 0.85$} & \multicolumn{1}{l|}{$0.62 \pm 0.61$} & \multicolumn{1}{l|}{$1.12 \pm 0.80$} & \multicolumn{1}{l|}{$0.51 \pm 0.39$} & \multicolumn{1}{l|}{$0.41 \pm 0.27$} & \multicolumn{1}{l}{$0.31 \pm 0.22$} \\ \cline{2-9} 
 & CNN & \multicolumn{1}{l|}{$1.43 \pm 0.64$} & \multicolumn{1}{l|}{$0.59 \pm 0.49$} & \multicolumn{1}{l|}{$0.69 \pm 0.49$} & \multicolumn{1}{l|}{$0.88 \pm 0.60$} & \multicolumn{1}{l|}{$0.61 \pm 0.48$} & \multicolumn{1}{l|}{$0.56 \pm 0.45$} & \multicolumn{1}{l}{$0.49 \pm 0.39$} \\ \hline
\end{tabular}
}
\end{table}
\vspace{-1em}

\section{Discussion and Conclusion}
In this work, we have presented a novel strategy to enable learning a deep metric for multimodal image registration from substantially misaligned data. Specifically, symmetrization and dithering are proposed to reduce bias and effectively merge multiple modes in the response function that are caused by the distribution of the misregistration in the training data. Although bigger misalignment can be captured by using larger patches, having limited memory capacity prevents us from increasing the patch size. To overcome this, an online data generation and training can be performed which drastically will increase the training time. To address this issue, we proposed a multi-resolution multi-shot approach. Currently, this approach comes with a mild interpolation artifact in the early-state response functions that can be seen in Figure \ref{fig:activations}d, it is not evident in the final stage response functions.

We demonstrated that our learned deep metric can effectively be used for the task of rigid registration and significantly outperform MI in estimating translation parameters. We believe applying dithering in rotation, in addition to the translation, will increase the performance with respect to rotation.  

In future work, we plan to expand our framework into nonrigid registration and evaluate it on a more difficult registration task: ultrasound to MRI registration, with resection. 
In conclusion, data augmentation via symmetrization and  dithering is an effective strategy that discharges the need for well-aligned training data -- this brings deep metric registration from the realm of supervised to semi-supervised machine learning.


\begin{thebibliography}{99}

\bibitem{simonovsky2016deep} Simonovsky, M., Guti´errez-Becker, B., Mateus, D., Navab, N., Komodakis, N.: A
deep metric for multimodal registration. In: MICCAI. (2016) pp. 10-18

\bibitem{cheng2016deep} Cheng, X., Zhang, L., Zheng, Y.: Deep similarity learning for multimodal medical
images. CMBBE (2016) 1-5

\bibitem{viola1997alignment} Viola, P., Wells III, W.M.: Alignment by maximization of mutual information.
International journal of computer vision 24(2) (1997) 137-154

\bibitem{Litjens2017ASO} Litjens, G.J.S., Kooi, T., Bejnordi, B.E., Setio, A.A.A., Ciompi, F., Ghafoorian,
M., van der Laak, J., van Ginneken, B., S´anchez, C.I.: A survey on deep learning
in medical image analysis. Medical image analysis 42 (2017) 60-88

\bibitem{Wu2013UnsupervisedDF} Wu, G., Kim, M., Wang, Q., Gao, Y., Liao, S., Shen, D.: Unsupervised deep feature
learning for deformable registration of mr brain images. In: MICCAI. (2013) 649-	656

\bibitem{Sokooti2017NonrigidIR} Sokooti, H., de Vos, B.D., Berendsen, F.F., Lelieveldt, B.P.F., Isgum, I., Staring,
M.: Nonrigid image registration using multi-scale 3d convolutional neural networks.
In: MICCAI. (2017)

\bibitem{Miao2016ACR} Miao, S., Wang, Z.J., Liao, R.: A cnn regression approach for real-time 2d/3d
registration. IEEE Transactions on Medical Imaging 35 (2016) 1352-1363

\bibitem{powel} Powell, M.J.D.: An efficient method for finding the minimum of a function of several
variables without calculating derivatives. The Computer Journal 7(2) (1964) 155-162

\bibitem{650848} Thevenaz, P., Ruttimann, U.E., Unser, M.: A pyramid approach to subpixel registration based on intensity. IEEE TMI 7(1) (Jan 1998) 27-41

\bibitem{1088973} Schuchman, L.: Dither signals and their effect on quantization noise. IEEE Transactions on Communication Technology 12(4) (December 1964) 162-165

\bibitem{ZagoruykoK15} Zagoruyko, S., Komodakis, N.: Learning to compare image patches via convolutional neural networks. In: 2015 IEEE (CVPR). (June 2015) 4353-4361

\bibitem{Kingma2014AdamAM} Kingma, D.P., Ba, J.: Adam: A method for stochastic optimization. CoRR (2014)

\bibitem{Srivastava2014DropoutAS} Srivastava, N., Hinton, G.E., Krizhevsky, A., Sutskever, I., Salakhutdinov, R.:
Dropout: a simple way to prevent neural networks from overfitting. Journal of
Machine Learning Research 15 (2014) 1929-1958

\bibitem{BrainDev33:online} IXI: Information eXtraction from Images. http://brain-development.org/

\end{thebibliography}

\end{document}